# A PyTorch-Enabled Tool for Synthetic Event Camera Data Generation and Algorithm Development

Joseph L. Greene[1*], Adrish Kar[2], Ignacio Galindo[2], Elijah Quiles[1], Elliott Chen[2], Matthew Anderson[1]

[1] Georgia Tech Research Institute, Electro-Optical Systems Lab, Atlanta, GA, 30332, USA

[2] Georgia Tech, College of Computing, Atlanta, GA, 30332, USA

*Joseph.Greene@gtri.gatech.edu; phone:1 404.407.7162

## ABSTRACT

Event, or neuromorphic cameras, offer a novel encoding of natural scenes by asynchronously reporting significant changes in brightness, known as events, with improved dynamic range, temporal resolution, and lower data bandwidth when compared to conventional cameras. However, their adoption in domain-specific research tasks is hindered in part by limited commercial availability, lack of existing datasets, and challenges related to predicting the impact of their nonlinear optical encoding, unique noise models, and tensor-based data processing requirements. To address these challenges, we introduce Synthetic Events for Neural Processing and Integration (SENPI) in Python, a PyTorch-based library for simulating and processing event camera data. SENPI includes a differentiable digital twin that converts intensity-based data into event representations, allowing for evaluation of event camera performance while handling the non-smooth and nonlinear nature of the forward model. The library also supports modules for event-based I/O, manipulation, filtering, and visualization, creating efficient and scalable workflows for both synthetic and real event-based data. We demonstrate SENPI's ability to produce realistic event-based data by comparing synthetic outputs to real event camera data and use these results to draw conclusions on the properties and utility of event-based perception. Additionally, we showcase SENPI's use in exploring event camera behavior under varying noise conditions and optimizing event contrast thresholds for improved encoding under target conditions. Ultimately, SENPI aims to lower the barrier to entry for researchers by providing an accessible tool for event data generation and algorithmic development, making it a valuable resource for advancing research in neuromorphic vision systems.

## 1. INTRODUCTION

Event-based, or neuromorphic, cameras adapt how imaging platforms encode information from the natural world through a bio-inspired paradigm that records binary indicators of temporal changes in the log intensity, known as the brightness, through timestamped tensors in a low bandwidth and asynchronous data stream [1], [2]. Through this novel architecture, these cameras support a >120dB dynamic range and >1-10kHz effective framerates in a compact and low-cost formfactor [3], [4]. In addition, their intrinsic data representation prioritizes temporally rich features while minimizing blurring and redundancy between frames [5], [6]. Encouraged by these properties, event cameras have found increasing success as ultrafast machine vision tools across scales and domains, such as in fault detection [7], automotive vision [8], optical flow [9], microscopy [10] as well as space and situational awareness [11]. However, as these cameras continue to find promise as alternative imaging sensors, their adoption creates additional attention towards overcoming the unique challenges in utilizing this technology. In particular, event-based technologies are in part constrained by their limited commercial availability and lack of existing datasets alongside their nonlinear optical forward model, ill-characterized noise models and tensor-based readout scheme [2], [5]. These traits introduce barriers to entry when innovating on existing work, such as in predicting performance in new applications or developing data-driven algorithms to push state-of-the-art performance.

To partially alleviate the above challenges, recent work has developed novel event camera simulators which seek to generate realistic event-based data through direct video-to-event conversions [12], [13], physics-guided simulation [14] or neural surrogates [15], [16], [17] to augment algorithmic performance and explore behavior. This line-of-research is particularly important as event-based cameras introduce camera-specific properties that must be tuned, such as contrast thresholds, that dictate their noise sensitivity and event-encoding fidelity [18]. In addition, other efforts have noted that properly representing physical phenomena such as low-light behavior [19], latency [18] as well as noise effects [20] is essential to achieving accurate synthetic data to inform data-driven solutions [21] and domain adaptation [22]. While this research direction addresses uncertainty in performance and augments the limited available data, it does not alleviate the challenges associate with the event data representation. Unlike conventional frame-based operation, event-based technology uses an asynchronous pixel readout that populates a low bandwidth data stream with 4D tensors containing the timestamp, x, y pixel array position and the polarity of the detected event [23]. When compared to intensity-based frames, this event-based data reduces redundancy and bandwidth requirements but trades structure and context typically leveraged by traditional image processing algorithms to enable vision-based tasks [24]. To overcome this challenge, recent applications invoke a number of strategies. In domain-specific tasks, researchers may apply physical priors such as motion assumptions (such as in optical flow [25] or tracking [26]), weighted temporal proximity (such as with time surfaces [27], [28], [29]) or sensor geometry (such as in multi-sensor configurations [30]) to convert raw events into representative patches, frames or full spatiotemporal volumes. In cases where the desired physical transform is ill-defined, such as in events-to-intensity transforms, other efforts use modified deep learning architectures leveraging custom datasets [31], [32] as well as unsupervised [9] and self-supervised [33], [34] methods. In an effort to inject structure back into the event feed, many solutions choose to reshape the tensor-based event stream into event frames or spatiotemporal volumes at the cost of additional latency to perform these operations. Despite this trade-off, seeking these structured representations offers value as the higher dimensional event representations enables inherent denoising by leveraging spatiotemporal continuity [35], [36] as well as the direct application of traditional frame-based image processing algorithms on the output [37]. Balancing the strengths of traditional and event-based imagers, recent effort has identified that fusing intensity-based and raw tensor-based event-based data might prove promising as it may leverage the context and structure of traditional frames with the low latency and temporally rich features of tensor-based event data to push speed, accuracy and utility of imaging systems [24].

The result of the above discussion is that proper integration of event camera into novel research directions is non-standard and, depending on the application, leverages domain-specific information and processing to enable success. As such, it is clear that the field would benefit from a general tool that could help predict event-based performance in new environments as well as unify event camera system design and algorithmic development to rapidly innovate new end-to-end pipelines. To meet this broad need, the tool should be general, scalable and support an accurate physics-informed simulation, recent advancements in event data processing as well as deep learning development. To meet this need, we introduce a custom PyTorch-enabled library entitled Synthetic Events for Neural Processing and Integration (SENPI) in Python. This tool is designed to build on existing photometric simulators to convert input radiometric information into realistic event-based data while supporting downstream processing. As a whole tool incorporates inspiration from prior work to offer the following novel capacities:

1. Custom implementation of an event-based camera forward model considering sensor parameters and noise to produce high-fidelity pseudo-event tensor data as well as event-based surrogate frames for training algorithms.

2. A modular framework designed to support targeted needs such as event conversion, I/O, filtering, manipulation, and visualization. It draws inspiration from literature to provide building blocks for developing event-processing pipelines for real and synthetic data.

3. Implementation in Python / PyTorch to enable a scalable and modular framework that may utilize GPU-acceleration and intrinsic differentiability to directly integrate and optimize physical parameters alongside downstream neural networking in deep optics routines.

We show an overview of the SENPI suite in **Fig. 1**. We believe that SENPI will find broad success in supporting fundamental studies, predictive analysis, data augmentation, event-based algorithmic development and neural network development. By supporting a modular design, we also believe that researchers will find the tool easy to implement and modify for custom applications. First, we discuss the design and functionality of SENPI in **Sec. 2**, focusing on the implementation of the nonlinear and non-smooth (i.e. discontinuous) event-based forward model, alongside the development of a corresponding differentiable backward method. Further information on the implementation of the event-based forward model and the incorporation of noise sources can be found in **Appendix A**. We verify SENPI's utility by comparing the quality of synthetic data and presenting a novel study that shows the preliminary optimization of event-based parameters under noisy conditions in **Sec. 3**. Overall, we believe this work will offer novel capabilities for designing and deploying event-based systems and algorithms, as well as highlight the utility of high-fidelity digital twins in advancing event-based applications.

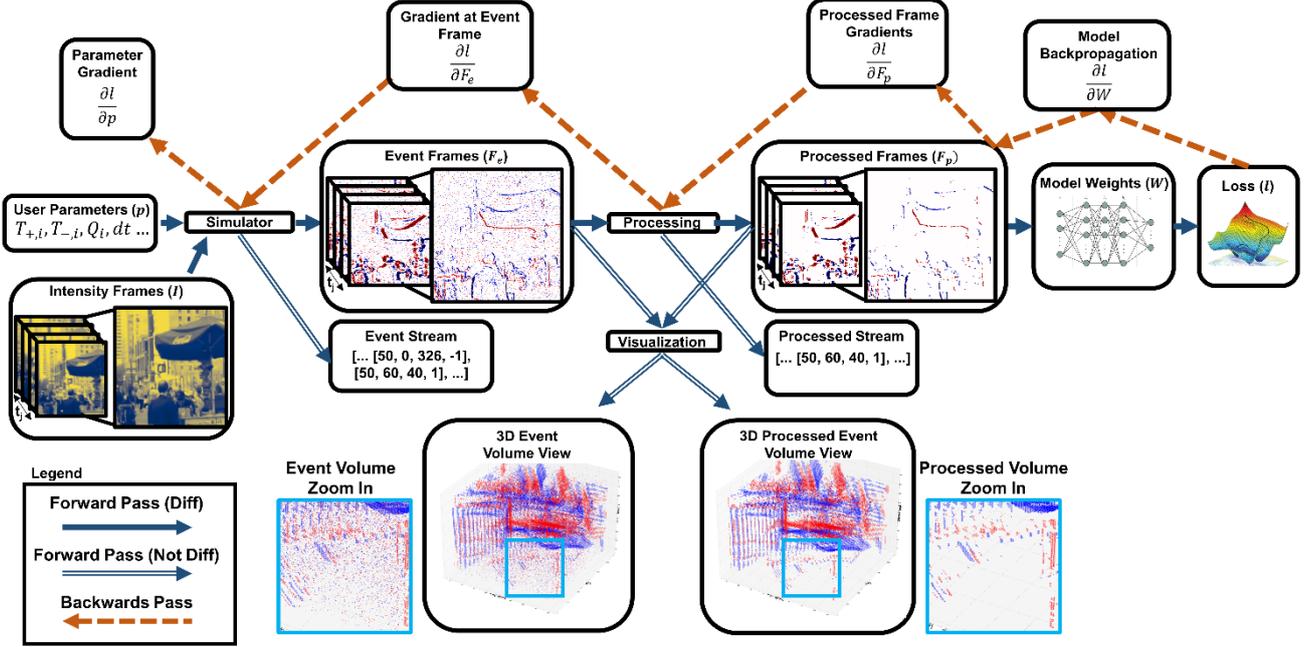

**Fig. 1: Overview of SENPI.** SENPI is a modular event-simulation suite built in PyTorch. This suite is designed to transform photometric frames into event-based data and enable downstream processing through first-order approaches or subsequent neural processing. By supporting a differentiable event-frame branch, SENPI may optimize physical alongside neural parameters simultaneously to create novel event-based architectures alongside producing high-fidelity surrogate data for domain-specific tasks.

## 2. METHODOLOGY

Here, we utilize the forward model described in **Appendix A** to develop the SENPI event-based simulator as well as explore recent work in event data manipulation, filtering and visualization to develop the algorithmic development suite. To ensure generality and maximize utility in Python/PyTorch, SENPI upholds three principles: (1) a scalable framework with modular submodules for intuitive event-based workflows, (2) an organized implementation allowing future researchers to customize and innovate as needed, and (3) support for common event data types and Python structures to ensure broad accessibility. Here, SENPI is designed to operate on Numpy arrays, Pandas data frames as well as PyTorch tensors with the ability to manipulate, filter and visualize each as well as convert between types. In addition, the functionality of PyTorch tensors shall preserve the key benefits of tensor-based processing, primarily GPU acceleration and auto-differentiation. To clarify terminology, moving forward "PyTorch tensors" will refer to the native data structure of the PyTorch library while "event tensors" are event-based data in the form presented in **Eq. A10**, which may be represented digitally in a variety of Python data structures. We highlight the key contribution of SENPI as well as their novel implementation in the following sections.

2.1. Event Model Simulation

The event-based simulator in the senpi.sim submodule is a standalone class that converts intensity frames into event-based data. It uses user-defined parameters to specify camera properties like temporal resolution, refractory period, thresholds, and noise parameters (hot pixels, shot, dark, leak noise, and pixel-wise contrast deviations). Shot noise is model through a Poisson node on the photometric input data, dark noise is model through an additional per-pixel buildup of photoelectrons within an integration period and leak noise is model as a uniform random probability that a "off" pixel (i.e. pixel that did not trigger an event) generates an erroneous positive or negative event. Similar to previous simulators, we also enforce a maximum voltage before saturation, here captured through a user-defined pixel well capacity, to normalize intermediate voltage values between 0 and 1 [13]. To aid in customizing SENPI, recent work suggests quick and effective methodology for calibrating event-cameras to characterize these quantities on a per-camera basis to reduce model mismatch when simulating off-the-shelf solutions [38], [39].

After defining user parameters, the simulator utilizes the forward model described by **Eqs. A1-A9** to generate synthetic event-based data. Since the input supports frames, SENPI implements a frame-based forward model where indexing is used to implement the per-pixel updates. This choice allows for this class to store all pixel-wise quantities, such as leak rate or threshold, as 2D maps in the internal attributes of the simulator class. Once generated, this simulator may return the event-based data as either frames and/or 4D event tensors described by **Eq. A10**. The core philosophy of this decision is twofold: 1.) reshaping and annotating the final event tensors is slow leading to artificial overhead and 2.) using frame-based operations on event-based data is still a rich area of research we wish to support. In an effort to remain general, this initial version of SENPI neglects a few architecture specific behaviors, such as the influence of light level on latency, and in-the-loop event filtering that might arise in other simulators [18]. However, we recognize that our implementation can be easily modified to reflect specific camera behaviors.

One key innovation is in preserving differentiability in the forward simulation. A growing area of optics research is in deep optics where "physical" optical parameters and "digital" network parameters are co-optimized to enable next generation computational imaging platforms [40], [41], [42]. However, this strategy demands that the optical simulation remains differentiable [43], which proves challenging as event-based models include singularities (e.g. log) as well as non-smooth indicator functions. To address this challenge, we implement two strategies. Firstly, we recognize that event-based cameras should exhibit a minimum response to 1 photon due to the quantized nature of light and produce no current under dark conditions. To address this observation, we develop a custom log class that handles the unrealistic singularity with a forward method that returns 0 for all inputs less than 1 and returns a gradient of 0 on the backwards methods. Next, we relax the bound of indicator functions using a sigmoid approximation [44], [45], which exhibits a tunable steepness that controls the accuracy when compared to the original indicator function vs the instability in the exponential term [46]. In the context of event-based detection, the steepness of the sigmoid corresponds to the leakiness surrounding spatial events which allows the network to preserve information from the underlying scene in the backpropagation. To balance the need to remain accurate with the need for differentiability, we deploy a forward-backward methodology, where the soft approximation is upheld in the backward calculation, similar to other non-convex algorithms that require relaxation [47], [48]. Another key challenge in gradient preservation is the influence of stochastic nodes, such as the introduction of shot, dark and leak noise sources. Fortunately, leak noise may be handled through reparameterization [49], dark noise is not signal-dependent and off the critical path of backpropagation, and shot noise may be handled through a score gradient method [41]. Altogether, these design practices create an event camera simulator that balances accuracy, speed, and differentiability, enabling the rapid production of synthetic event-based data to predict performance and train or augment deep learning models. We showcase a full graphical representation of the forward and backward SENPI pass in **Fig. 2**.

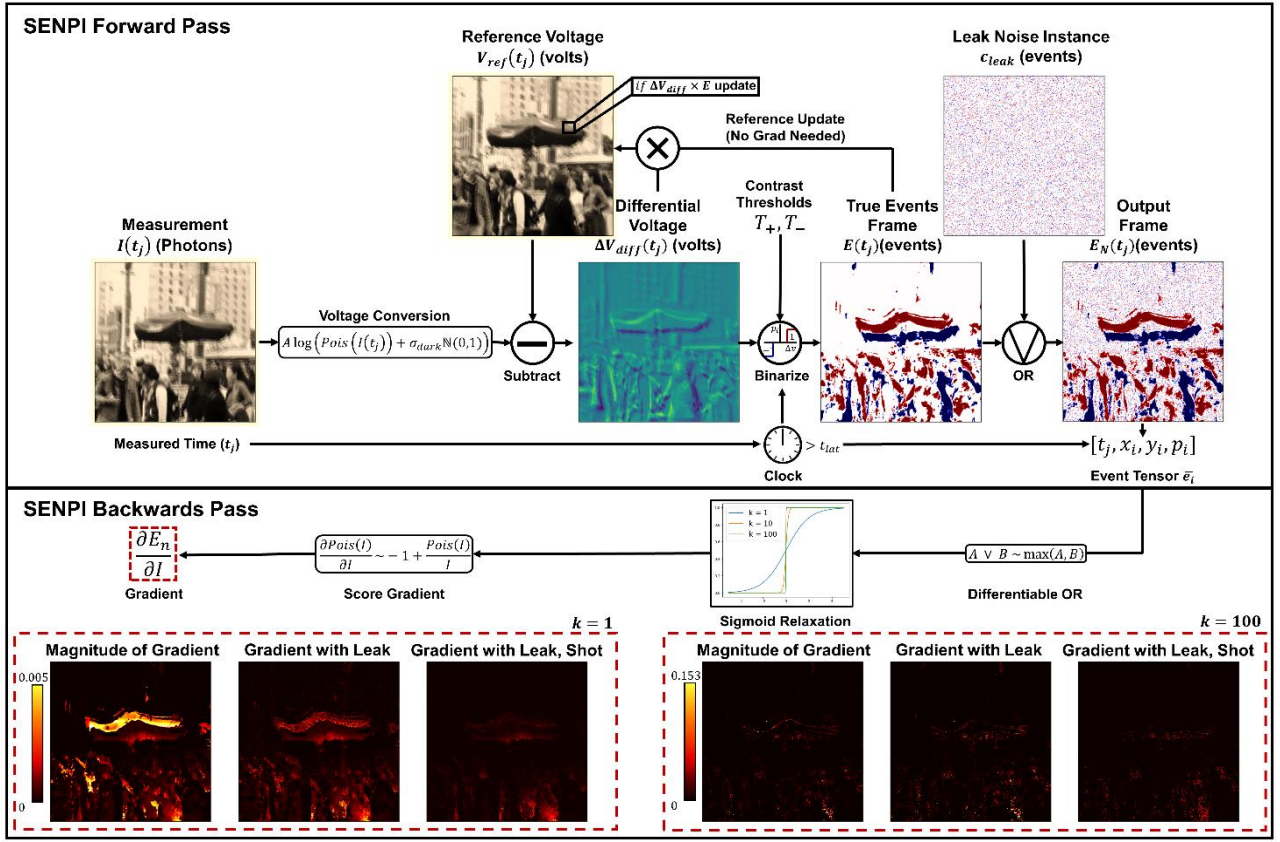

**Fig. 2: SENPI forward and backwards model.** Flowchart of the SENPI forward (top) and backwards (bottom) model emphasizing the operations utilized to achieve realistic behavior in the forward path and their associate relaxation in the backwards pass. The backwards pass of stochastic nodes is alleviated through reparameterization and indicator functions are alleviated through a sigmoid approximation. As demonstrated in the red boxed inlets of the backwards pass, the user may control the sharpness of the sigmoid approximation which trades the gradient magnitude with the proximity of relevant information when compared to the event-based output frame as shown under noiseless and noisy conditions.

2.2. Data Manipulation

The SENPI library provides a suite of data manipulation tools to streamline event-based data processing. Each manipulation operation is organized into a class structure with dedicated methods to process the supported data types, consisting of PyTorch tensors, Numpy arrays and Pandas DataFrames. These data types are selected due to their high level of support and utility in organizing and processing data within Python and SENPI natively supports the capability to convert between data types at any stage of the processing pipeline. Primarily, we intend for SENPI to be utilized alongside PyTorch tensors for intermediate processing to enable users to leverage GPU acceleration and PyTorch tensor-based operations for efficient and scalable workflows. To balance the need to generate properly structured event-based data alongside promote the accessibility of frame-based processing, SENPI's manipulation operations are designed to operate on either 1D event tensor-based and 2D event frame-based structures. We contextual the unique benefits, challenges and implementation of each choice below.

2.2.1. Event Tensor-based Filtering

Event tensor-based filtering in SENPI manipulates raw data in its native tensor format, offering a compact view of temporal changes. It also improves memory efficiency by rejecting null pixels. Based on recent work, SENPI chooses to support relevant first-order statistical filters due to their ability to remove noise events from the raw event stream by highlighting salient dynamic event patterns without the need to learn latent data distributions. The input of each filter method is expected to be an event stream in either a PyTorch tensor, Numpy structured array, or Pandas data frame. Note that, as of right now, the event tensor-based filtering methods are not differentiable due to the slicing operations involved in the filtering process. Using this common structure, we implement various filtering techniques from literature with different applied filtering philosophies and prior assumptions, including:

- **Polarity Filter:** This basic filter class allows selective filtering of events based on their polarity. Users can isolate positive or negative events through the use of this class.
- **Background Activity Filter (BAF):** SENPI supports BAF, which is a type of filter that rejects events with inadequate spatiotemporal support as likely random and sparse noise events. The filter operates using a timestamp map, a $W \times H \times 2$ tensor where $W$ and $H$ represent the width and height of the event frame and the third dimension separates timestamps for negative and positive polarity events. For each pixel, the timestamp map records the most recent event timestamps, enabling spatiotemporal comparisons [50].
- **Inceptive Events Filter (IEF):** SENPI also supports IEF, a filter designed to preserve indicative features in the event stream, such as corners and edges, that are essential for object detection and classification. This filter is parameterized by two temporal proximity thresholds, $t_{IEF}^-$ and $t_{IEF}^+$. Here, IEF looks at a given pixel over all time and keeps an event there exists a successive event of the same polarity within the intervals $[t_j - t_{IEF}^-, t_j]$ and $[t_j, t_j + t_{IEF}^+]$ [51]. SENPI also implements a polarity-agnostic IEF, which keeps an event if there is another event of any polarity instead.
- **YNoise Filter:** YNoise is a recent filtering technique that comprises of two subsequent filtering operations [52]. The first phase involves filtering out random noise via coarse filtering and the second phase is for fine filtering, with an emphasis on rejecting hot pixels.

2.2.2. Frame-based Filtering

To improve the applicability of the SENPI, we additionally support frame-based filtering, which modifies the functionality above to apply on structured event frames instead of an unstructured event stream. We take care to create novel implementations of the above filters using reshaping, convolution and masking operations to preserve auto-differentiation and further push acceleration on PyTorch tensors. We believe this filtering modality will couple with the option for frame-based event data from SENPI to offer a more structured and accessible pipeline for designing, testing and optimize event-based technology in deep learning-based or computer vision-based tasks. To standardize the event frame data in a familiar format for array-based processing, we organize the event-based data into [N, C, X, Y] arrays, where N is the number of frames where each frame corresponds to a unique timestamp worth of information, C is the number of channels which is natively 1 for greyscale event-based data and X, Y are spatial dimensions. Pixels that do not contain an event within the time interval are set to 0.

2.3. Data Visualization

Data visualization is essential for interpreting event-based data. As such, SENPI includes tools to represent event data spatially and temporally to aid researchers in interpreting complex event stream dynamics. These visualization methods include: reconstructed 2D intensity frames, 3D spatiotemporal volumes, and time surfaces, each offering a unique perspective on event-based data.

One method of visualization is to reconstruct of 2D intensity frames from the underlying event stream. This method applies a continuous-time intensity estimation approach similar to that described in [53],where events update an image state asynchronously based on specified contrast thresholds and temporal decay. SENPI supports both event-driven and frame-rate-based generation modes, offering options for spatial smoothing through Gaussian or bilateral filtering. However, we recognize that more recent algorithmic works may be integrated and provide more accurate intensity-based estimates [54], [55]. Beyond frame-based representations, SENPI also enables 3D visualization of event data point clouds where events are color-coded based on polarity. In addition, SENPI supports the utilization of visualization techniques known as time surfaces, which weight a spatial neighborhood by recent time history of events to provide spatiotemporal context to visualizing events. They can be visualized as either 2D heatmaps, which emphasize event recency through color intensity, or 3D surface plots, which provide a detailed representation of temporal activity across spatial dimensions. These surfaces are favorable due to their ability to emphasize persistent structures in event streams and aid in the tasks of representation, filtering, and feature extraction and form a cornerstone of visualization in SENPI. In an effort to push the visualization of useful events, all time surfaces are implemented in polarity sensitive or polarity agonistic implementations, which preserve or ignore polarity information respectively before plotting. In addition, all time surfaces may be constrained to a local spatial window to emphasize areas of significant action. SENPI also supports several basic time surface modes that can be modified to create better input feature descriptors for models as in [55]:

- **Exponential Time Surface:** The exponential time surface temporally weights events within a local neighborhood using a user-parameterized time decay constant to emphasize spatio-temporal continuity. This representation emphasizes more recent events, diminishing the contribution of older events exponentially over time [27]. This representation is often used in conjunction with spatial aggregation techniques, such as Histograms of Averaged Time Surfaces (HATS), to enhance the descriptive power of event stream features and perform downstream tasks such as object detection, filtering, and reconstruction [10], [27].
- **Count:** The count time surface mode simply tallies the number of events at each pixel position within the batch, irrespective of their temporal order or polarity. This produces a representation highlighting spatial activity intensity.
- **Average:** The average mode computes the mean of event polarities at each pixel, providing a polarity-weighted view of the spatial activity. This mode can serve as a simple form of polarity-aware spatial encoding.
- **Average Absolute:** Similar to the average mode, this computes the mean of the absolute values of event polarities, effectively ignoring the direction of intensity changes but still capturing activity magnitude.

## 3. RESULTS

3.1. Understanding and Optimizing an Event-based Camera under Noisy Conditions using SENPI

Event-based technology, as a novel imaging tool, presents several open questions regarding its properties, primarily due to its nonlinear forward model. These challenges are further compounded by the limited availability and the relatively few existing studies. We propose that high-fidelity digital twins, such as those supported by SENPI, can facilitate targeted simulations to advance research by developing preliminary theories and testing hypotheses while the field works to validate the novel behavior of event-based technologies through experimentation. We also argue that digital twins provide a unique opportunity to gain valuable insights by allowing us to control physical phenomena in non-physical ways, such as tuning or removing noise sources to deepen our understanding. This section uses SENPI to develop and test novel event camera models related to the encoding of noise and ideal impulses as a function of camera parameters and incident intensity. Our goal of this study is to propose an information-driven methodology using SENPI to optimizing event-based parameters to maximize the encoding of reliable information under specific conditions before data collection. We use this analysis to additionally verify novel theory on the event sensitivity threshold, or the theoretical amount a signal must change to trigger an event, as introduced in **Appendix B**.

To frame this study, we first simulate how noise affects event camera output on a temporally static but spatially varying background, representing the mean number of photons captured between time points, as shown in **Fig. 3A (top)**. Under fixed conditions (contrast thresholds = 0.01, no latency), we simulate the resultant output between 10000 successive time points to determine the probability of triggering a noise event, here referred to as the "false alarm rate", under different noise considerations (e.g. shot noise, dark noise and leak noise) in an ideal case with high sensitivity, as shown in **Fig. 3B**. We build on this baseline case by introducing a 100-photon impulse on top of the static background and analyzing how noise impacts our ability to detect it as shown in as shown in **Fig. 3A (2nd, 3rd, and 4th rows)**. Here, we find that noise introduces statistical fluctuations in the resultant signal, leading to a decreased chance of detecting the impulse under different background levels, as shown in **Fig. 3C**. Notably, the probability of detecting the impulse decreases to a minimum value of 0.5 at the event sensitivity threshold, as expected, because we assume that the noise sources have a zero mean (see **Appendix B**).

Next, we examine how adjusting event camera parameters affects performance in order to determine the optimal settings. Specifically, we simulate how latency (see **Fig. 3D**) and contrast thresholds (see **Fig. 3E**) influence the false alarm rate. As shown in Fig. 3D, latency scales the false alarm rate by a factor of $\Delta t/R$, where R represents the refractory period.

Introducing latency effectively 'turns off' a pixel after a noise event occurs, but it does not alter the underlying noise statistics. While this suggests that high latency is beneficial, the danger is that no events, including true signal events, may be detected in this downtime. As such, it is recommended to minimize latency in an event camera and use other parameters, such as contrast threshold, to reduce noise. By **Fig. 3E**, the impact of contrast threshold additionally impacts the alarm rate but in a less straightforward relationship. Again, this infers that high contrast thresholds are beneficial; however, increasing the contrast threshold increases the event sensitivity threshold and places higher demands on the amount a signal must change before we can register an event.

To explore a methodology to set the contrast threshold, we first update our model to predict the appropriate amount of background photons we may expect to collect in an event-based imaging system through diffuse solar radiation in **Fig. 3F**. We identify two periods of operation we wish to optimize performance over: low background morning hours and high background daytime hours. Next, we visualize the trade-off between probability of detection and false alarm rate under variable background levels in representative receiver operation characteristic (ROC) curves, shown in **Fig. 3G**. We may integrate the ROC curves for the relevant background levels to determine the area under the curve (AUC) as a function of impulse strength and contrast threshold. Under this methodology, AUC is a standard method for determining the effectiveness of a classification system under noise, where a value of 1 indicates perfect classification and a value of 0.5 or lower indicates a coin-flip or poorer classification. As such, we may use the resultant graphs at our two operating conditions (morning, see **Fig. 3H** and day, see **Fig. 3I**) to determine the best contrast threshold for a given impulse strength as well as a corresponding value for the reliability of the resultant event stream. From these plots, we identify three regimes of interest. At low contrast thresholds, the event camera is most sensitive, leading to better encoding of impulses of varying strength, but it also generates the highest amount of noise. At high contrast thresholds, the event camera becomes less sensitive to noise may place too rigorous demands on the required impulse to trigger an event, especially at high background levels (see **Fig. 3I**), leading to poor event encoding. Between the two, we can find values for contrast thresholds and assumed impulse strength that optimally balance both.

Overall, this novel analysis showcases how SENPI may be used to study fundamental noise sensitives in event-based vision and characterize how well an event camera may work as an encoding tool as a function of signal, noise, camera and background parameters. One strength of this methodology is its ability to pre-optimize event-based performance for target tasks without requiring data, relying only on information about signal and background levels. We believe that using digital twins to simulate event camera parameters will continue to play a significant role in determining optimal performance requirements for these novel tools to enable the practical use of event-based cameras in real-world applications.

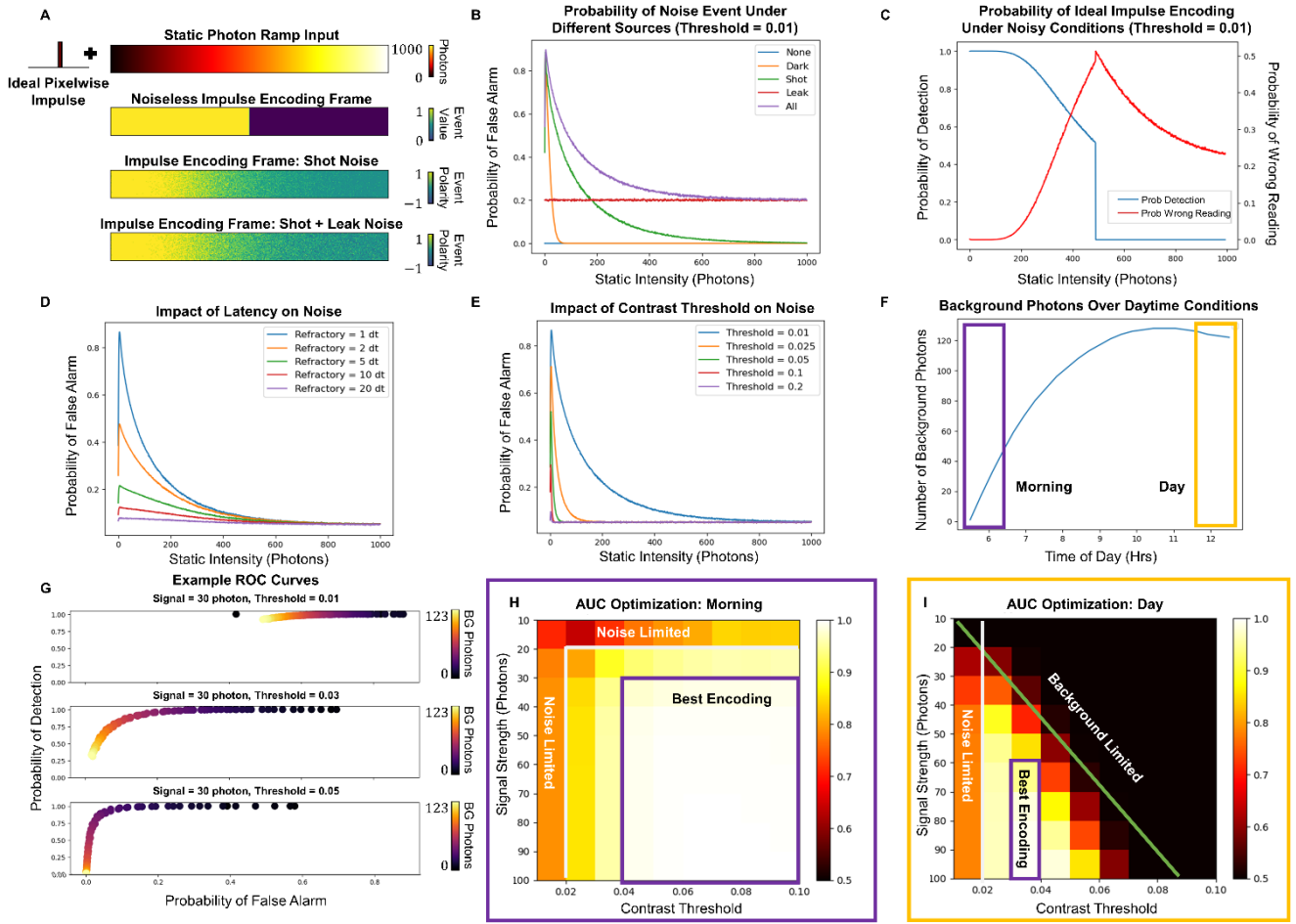

**Fig. 3: Impulse Encoding of an Event-Based Camera.** (**A**) Encoding of an ideal impulse under noiseless and noisy conditions. (**B**) Monte Carlo determination of an event camera false alarm (e.g. noise) rate due to different assumed sources on a static background. (**C**) Probability of encoding an impulse vs achieving an incorrect measurement under noisy conditions. (**D**) Influence of latency on the false alarm rate. (**E**) Influence of contrast thresholds on the false alarm rate. (**F**) Representative number of background photons captured by an optical system at event framerates. (**G**) Example ROC curves indicating noise-fidelity trade-off as a function of background strength and threshold. (**H**) Determination of optimal signal strengths and thresholds for event-based encoding during simulated daytime operation. (**I**) Determination of optimal signal strengths and thresholds for event-based encoding during simulated daytime operation.

3.2. Characterizing SENPI Data Transformation on Natural Scenes

Finally, we seek to characterize how well the data generated through SENPI may be used to motivate solutions in machine vision applications. To confirm this utility, we seek to justify that SENPI data is structurally similar to true event-based data such that it may be used as a high-fidelity digital twin when exploring potential use-cases and when developing event-based algorithms. To motivate an answer, we first aggregate a database of natural images and split them into three datasets: a standard "intensity" set, a "simulated" set that SENPI converted from intensity- to event-based representations and a "true event" set that is comprised of open source data from real event cameras. The intensity-based data used in the "intensity" and "simulated" sets were aggregated across multiple open source and free-to-use videos while the true event-based dataset was collected from the Prophesee website and each set contains a diversity of objects ranging from people, animals, cars, and landscapes with static or moving cameras to represent the distribution of data one might expect across varied applications.

We perform principal component analysis (PCA) on each dataset and plot 300 randomly selected images in the reduced space. The role of PCA in this context is to highlight the presence of significant features irrespective of context, specific imaging sensor or noise to graphically represent the degree of similarity in the collected datasets. In this low-level representation, it is well known that standard intensity-based images tend to cluster based on the similarity of their features and that machine vision algorithms benefit from learning to recognize objects that exhibit large distances between their clusters as this indicates that their spatial profile is comprised of distinct features or patterns [56]. To emphasize this trend, we present representative images pulled from the PCA distribution (see **Fig. 4B, blue**) where each row comprise of the images corresponding to pairs of nearest neighbor points in PCA space and columns represent different clusters, as shown in **Fig. 4A**. Reinforcing this interpretation, the nearest neighboring points tend to arise from the same underlying video due to their high degree of similarity and different clusters represent different object categories ranging from a landscape (**Fig. 4A1-2**), cheetah pack (**Fig. 4A3-4**), and river (**Fig. 4A5-6**). Next, we look to the distribution of SENPI generated data (see **Fig. 4B, orange**) and true event data (see **Fig. 4B, green**) in this PCA space. Importantly, we note that SENPI data and true event data are closely and similarly clustered indicating strong similarity. Promisingly, we additionally comparing the power spectral density, which has proven to be a powerful metric when judging the quality of synthetic data for training neural networks [57], and find them to be in close agreement (see **Fig. 4C**). Altogether, we may conclude that despite arising from different natural sets and unknown camera properties, SENPI is able to convert intensity-based images into a representation that exhibits similar features and frequency composition as true event-based data showcasing its strength as a high-fidelity digital twin.

Plotting the SENPI and true event-based datasets in the PCA space additionally reveals an interesting trend. Neighboring points still tend to originate from the same video (see **Fig. 4D8-9, 10-11**), but the distance between clusters is significantly smaller compared to intensity-based data, with some outliers. Although these results may not be entirely surprising, they offer important insights from a machine vision perspective. Compared to intensity-based images, event-based imagery tends to reject low spatial frequency and low contrast information, which might contribute to the spatial representation of an object. Instead, it emphasizes features with high temporal significance between frames. As a result, the distance between event data clusters in PCA space is reduced as the extract features tend to follow similar spatial profiles (i.e. sharp edges surrounded by blank space). As a consequence, we may expect that frame-based image processing tasks, such as object classification, will not be as successful on event-based data as it lacks the subtle spatial cues to aid in distinguishing natural objects yet will prove superior in temporal-based tasks, such as object detection or tracking, as this data representation natively preserves temporal cues that exhibit high significance while rejecting low significance noise or clutter that might otherwise confound these algorithms. Additionally, we observe that the distance between an event-based data point and the origin of the PCA plot is more closely related to the number of triggered events rather than the object class. As a result, outliers tend to indicate the presence of a large fast-moving object (such as in the case of **Fig. 4D11**) or when the event camera itself is moving (such as in the case of **Fig. 4D12**) while points near the origin indicate more subtle motion (such as in the case of **Fig. 4D9-10**). Based on this trend, we propose that monitoring an event-based stream or video in a reduced space could serve as an alternative to traditional event-based image processing, allowing us to detect significant actions with minimal processing. As a whole, we believe that further studying the variations in event- versus intensity-based data will further assist in the understanding on how event-based technology changes our ability to observe the natural world and will serve as a basis for discovering additional use-cases that maximize the benefits of this emerging technology.

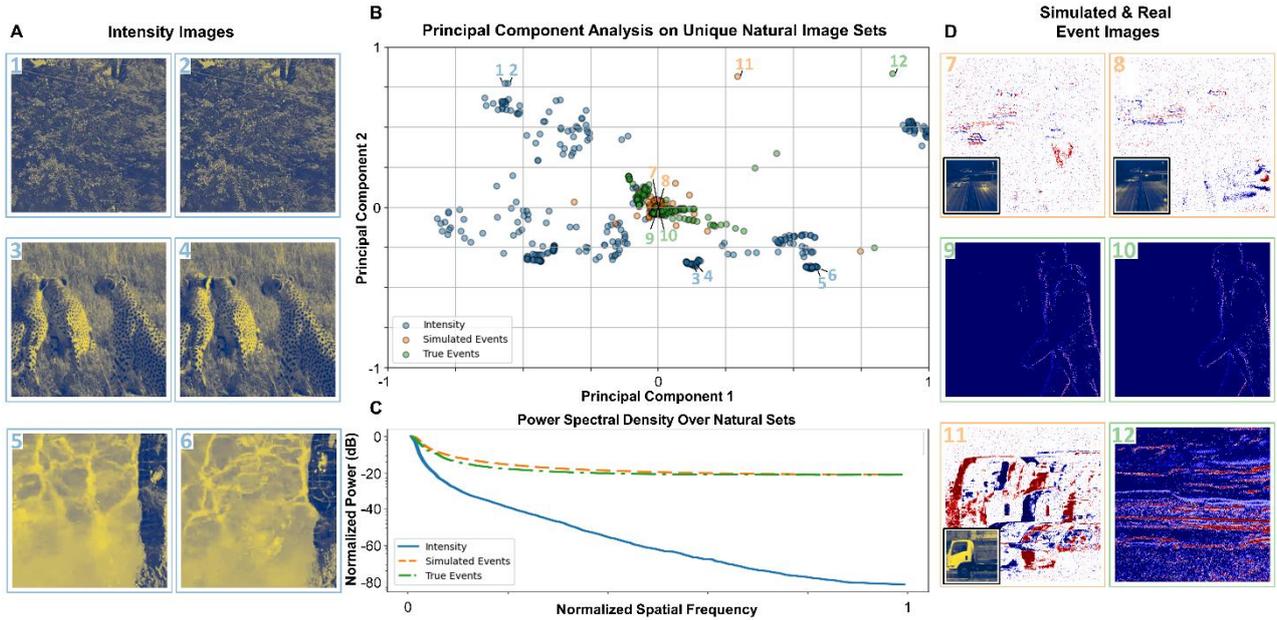

**Fig. 4: Characterizing SENPI as a Data Transformation Tool**. (**A**) First, we investigate the distribution of several different intensity-based images from diverse natural scenes in terms of their proximity in a reduced space. (**B**) Depiction of all intensity, SENPI and true event-based images in a reduced 2D space through PCA. (**C**) Comparison of intensity, SENPI and true event data in terms of their power spectral density over multiple natural sets. (**D**) Trends in SENPI and true event data in the reduced space.

## 4. CONCLUSION

In In conclusion, we introduce SENPI, a novel event camera simulation and data processing suite designed to advance physical studies, data augmentation, and algorithmic development. Implemented in PyTorch, SENPI provides a modular solution for integrating event-based simulation capabilities into existing photometric simulators. Our event-based simulator leverages a discretized version of the event-based forward model and employs relaxations of non-smooth operations to preserve differentiability. These innovations result in a GPU-accelerable and differentiable tool for generating synthetic event-based data. SENPI also includes an algorithm development suite with tools for event conversion, I/O, filtering, manipulation, and visualization while supporting various event data processes across popular data types. Using SENPI, we can study fundamental noise statistics in controlled simulations, gaining insights into how event-based technologies encode information and proposing an optimization routine to adjust event camera thresholds without requiring data acquisition. Furthermore, we demonstrate that SENPI accurately adapts real intensity data to an event-based representation, closely matching real event-based data across multiple analytic spaces.

In future work, we plan to further enhance SENPI's capabilities and explore event camera behavior in greater detail. One key limitation of any simulation tool is model mismatch, and we intend to conduct a comprehensive comparison between SENPI and off-the-shelf event-based cameras in controlled experiments. This process will help us refine our model and improve our understanding potential mismatch, such as event rejection during readout and architecture-specific event filtering. Additionally, we aim to identify and address overlooked noise sources and assumptions, such as offset noise and threshold distributions. Another area for improvement is in developing interpolation for event-based data. While existing strategies focus on whether events occurred in an interpolated interval, they neglect the reshuffling of events caused by latency. We believe addressing this challenge will allow SENPI to maintain high accuracy while operating on sparsely generated data. Overall, SENPI provides a high-fidelity digital environment to support the development of event-based technology. It offers essential building blocks for training, augmenting, and supporting event-based systems, while predicting performance and offering the ability to test novel theories. We believe this suite is a valuable tool for advancing machine vision tasks and fostering future innovations in event-based technology.

# CODE AVAILABILITY

The code for Synthetic Events for Neural Processing and Integration (SENPI) in Python is publicly available on GitHub at https://github.com/joeg18/senpi_ebi. The library is released under the MIT License, which permits free use, modification, and distribution, provided that the original authors are credited and the license text is included with any copies of the software. For further inquiries or any issues with the library, please feel free to open an issue on the GitHub page or contact the corresponding author at Joseph.Greene@gtri.gatech.edu.

# ACKNOWLEDGEMENTS

We would like to thank GTRI's Independent Research and Development (IRAD) funds for supporting this work. We additionally would like to thank Nathan Meraz, Jason Zutty, and Megan Birch for their constructive conservations and encouragement to pursue SENPI. On behalf of all authors, the corresponding author states that there is no conflict of interest.

# APPENDIX A: EVENT CAMERA MODELING AND DATA REPRESENTATION

Event-based pixels respond asynchronously and nonlinearly to changes in the incident intensity leading to deviations in how this technology is simulated when compared to a traditional camera. To balance accuracy with computational complexity, we look to previous work to incorporate a few standard assumptions: a.) the temporal variation in the signal is slow when compared to the framerate, b.) the event camera contrast threshold exhibit gaussian variability around a mean-value, c.) noise is dominated by leakage at the comparator but is influenced by shot and dark noise sources [2], [3], [18]. Currently, SENPI returns positive and negative events to match a Prophesee-style camera; however, this choice is not universal in the field. As such, while the follow analysis prioritizes this interpretation of event-based technology, the forward model is implemented in a manner such that it may be easily customized to match specific brands or technological advancements in the future. In addition, as simulation tool, this analysis will prioritize interpreting event camera behavior in discrete space and time representation to enable the generation of synthetic data on spatiotemporal grids.

To begin, a standard event-based pixel circuit consists of four main components: a photodiode, transimpedance transistor, amplifier and comparator. In brief, the role of the photodiode is to convert the incident photon flux into a photocurrent. The transimpedance transistor produces a voltage in response to the log of this photocurrent, which is subsequently amplified by the amplifier with a per-pixel gain. Finally, the comparator compares this instantaneous voltage to a stored reference voltage over time and outputs a binary indicator $\in [1, -1]$ when it exceeds a user-defined contrast. In practice, event-based pixels may incorporate additional elements to control intrinsic bandwidth, event signal filtering and latency [58]; however, we will ignore these effects for this preliminary analysis and for the sake of generality of this tool.

Here, we model a continuous domain photon flux $\Phi(x, y, t)$ is incident on the event camera sensor array with spatiotemporal coordinates $(x, y, t)$. The resulting photocurrent, $J(x, y, t)$ from the photodiode is:

$$J(x, y, t) = Q(x, y) \times \Phi(x, y, t) \times q_e \qquad \text{A1}$$

Where $Q(x, y)$ is the quantum efficiency and $q_e$ is the fundamental charge of an electron. To simply the notation, we introduce a spatial index, $i$, that represents the discrete linear positional index of an event camera pixel array and discretize the above quantities appropriately (such as integrating the continuous-space photon flux across the local pixel area). Under this representation, **Eq. A1** becomes:

$$J_i(t) = Q_i \times \Phi_i(t) \times q_e \qquad \text{A2}$$

Next, a subthreshold transistor scales its output in proportion with the log photocurrent and accumulates the resulting charge over the exposure time across a capacitor. This operation converts the continuous time photocurrent into a discrete time voltage with temporal index $j$ and binning associated with the exposure time, $\Delta t$, through the transform $t_j \sim t_0 + j\Delta t$, where $t_0$ is the continuous time start point (assumed to be zero unless otherwise specified). This discrete time voltage is subsequently amplified by an amplifier downstream. We may incorporate all amplification factors into a system gain, $A$, and describe the discrete time differential voltage at the comparator by:

$$\Delta V_{diff,i}(t_j) = A \int_{t_0+(j-1)\Delta t}^{t_0+j\Delta t} \frac{d(\log(J_i(t)))}{dt} dt \qquad \text{A3}$$

When the change in the photon flux is slow compared to the exposure time, we may approximate **Eq. A3** by:

$$\Delta V_{diff,i}(t_j) \sim V_i(t_j) - V_i(t_{j-1}) = A \log\left(J_i(t_j)\right) - A \log\left(J_i(t_{j-1})\right) \qquad \text{A4}$$

Where $V_i(t_j)$ is the instantaneous voltage accumulated at the comparator. Under this assumption, we may map the continuous time domain to the discrete time domain by interrogating the continuous time signal at fixed intervals instead of integrating over the interval. Physically, this interpretation enforces that the instantaneous photocurrent is reflective of the total photoelectron charge acuminated at the capacitor, which is a valid assumption at the small integration period utilized by event-based technology [3]. The comparator compares the differential voltage to user-defined but pixel-wise varying positive ($T_{+,i}$) and negative ($T_{-,i}$) contrast thresholds to determine a significant change and outputs the polarity of the change, $p_i(t_j)$ accordingly. As such, the polarity may be related by:

$$p_i(t_j) = \begin{cases} 1, & V_i(t_j) - V_i(t_{j-1}) \geq T_{+,i} \\ -1, & V_i(t_j) - V_i(t_{j-1}) \leq T_{-,i} \end{cases} \quad \text{A5}$$

When an *event is triggered*, the instantaneous voltage level at the current time point is saved to be used as a reference voltage at the subsequent time point. However, when an *event is not triggered*, the event pixel retains the reference voltage between timepoints. We accordingly modify **Eq. A4** to include a reference voltage variable, $V_{ref}(t)$ that incorporate this behavior:

$$p_i(t_j) = \begin{cases} 1, & V_i(t_j) - V_{ref,i}(t_{j-1}) \geq T_{+,i} \\ -1, & V_i(t_j) - V_{ref,i}(t_{j-1}) \leq T_{-,i} \end{cases} \quad \text{A6}$$

With the update rule:

$$V_{ref,i}(t_{j+1}) = \begin{cases} V_i(t_j), & \text{if } p_i(t_j) \\ V_{ref}(t_j), & \text{if not } p_i(t_j) \end{cases} \quad \text{A7}$$

Finally, event cameras do not generate events at every possible timepoint. In practice, event-based pixels are subjected to the effects of latency, which is a period of inactivity after triggering an event due to readout to the event stream and resetting of the comparator. To incorporate this behavior, we may introduce a new temporal variable, $\Delta t_{ref,i}$, which indicates the latency time of an event camera pixel and $t_{lat,i}(t_j)$, which tracks the inactivity period from the current time point after an event is triggered. We combine these new variables in an update rule:

$$t_{lat,i}(t_j) = \begin{cases} t_{lat,i}(t_{j-1}), & \text{if } t_{lat,i} > t_{j-1} \\ t_j, & \text{if not } p_i(t_{j-1}), t_{lat,i} \leq t_{j-1} \\ t_j + \Delta t_{lat,i}, & \text{if } p_i(t_{j-1}) \end{cases} \quad \text{A8}$$

We may combine the correction of **Eq. A6-8** into **Eq. A5** to present a final event camera pixel model of:

$$p_i(t_j) = \begin{cases} 1, & V_i(t_j) - V_{ref,i}(t_{j-1}) \geq T_{+,i} \ \& \ t_j \geq t_{lat,i}(t_j) \\ -1, & V_i(t_j) - V_{ref,i}(t_{j-1}) \leq T_{-,i} \ \& \ t_j \geq t_{lat,i}(t_j) \end{cases} \quad \text{A9}$$

Once the conditions of **Eq. A9** are met, the event information is output in the event stream as a 4D tensor containing the timestamp, pixel 2D indexing ($x_i, y_i$) and event polarity. Here, we model the event tensor, $\overline{e}_{ij}$ as:

$$\overline{e_{ij}} = [t_j, x_i, y_i, p_i(t_j)] \quad \text{A10}$$

Next, we aim to use SENPI to procedurally study the influence of noise on the output event stream under static input conditions. The first noise source we consider is the impact of shot noise, which may be included as a Poisson noise source that creates statistical fluctuation in the arrival of photons at the photodiode. We add the contribution of dark noise, which occurs due to photon-independent and thermally excited electrons that contributes to the resulting photocurrent as well as leak noise, which is erroneously generated events caused by leakage at the comparator. Here, we will assume that dark noise is described by a zero-mean Gaussian random variable parameterized by a variance, $\sigma_D^2$ and leak noise is described by a uniform random variable with leak chance, $c_{leak}$, to leak a positive or negative event. Altogether, we integrate the impact of shot and dark noise into **Eq. A4** as well as adjust the event generation condition to include leak noise in **Eq. A9**, yielding:

$$\Delta V_{diff,i}(t_j) \sim A \log(Pois(\Phi_i(t)) * q_e + \sigma_D \mathbb{N}(0,1)) - A \log(Pois(\Phi_i(t_{j-1})) * q_e + \sigma_D \mathbb{N}(0,1)) \quad \text{A4b}$$

Where $Pois(\ )$ is the Poisson operator on the input photon count at a given time interval and $\mathbb{N}(0,1)$ is a standard normal random variable and:

$$p_i(t_j) = \begin{cases} 1, & (V_i(t_j) - V_{ref,i}(t_{j-1}) \geq T_{+,i} \mid U(0,1) \geq 1 - c_{leak}) \ \& \ t_j \geq t_{lat,i}(t_j) \\ -1, & (V_i(t_j) - V_{ref,i}(t_{j-1}) \leq T_{-,i} \mid U(0,1) \leq c_{leak}) \ \& \ t_j \geq t_{lat,i}(t_j) \end{cases} \quad \text{A9b}$$

Where $U(0,1)$ is a standard uniform random variable.

### APPENDIX B: DERIVATION OF AN EVENT SENSITIVITY THRESHOLD

Unlike normal camera, which benefit from linear forward models and determinable thresholds of detection, the nonlinear nature of event-based technology introduces difficulties in predicting minimum detectable thresholds. As such, we seek to determine this "event sensitivity threshold" or the minimum detectable change in photons using the previous establish physical model. To begin, we plug **Eq. A2** into **Eq. A4**. and set the modified differential voltage equal to the positive and negative threshold in **Eq. A5**. We solve for the required photon flux at the current time point required to trigger a positive or negative event (i.e. $\Phi_i(t_j; p_i = 1), \Phi_i(t_j; p_i = -1)$), yielding:

$$\Phi_i(t_j; p_i = 1) = \Phi_i(t_{j-1}) \times \exp\frac{T_{+,i}}{A} \quad \text{A11}$$
$$\Phi_i(t_j; p_i = -1) = \Phi_i(t_{j-1}) \times \exp\frac{T_{-,i}}{A}$$

Next, we subtract $\Phi_i(t_{j-1})$ from each side of **Eq. A1** to determine the event sensitivity threshold, $\Delta\Phi_i(t_j, p_i = 1), \Delta\Phi_i(t_j, p_i = -1)$:

$$\Delta\Phi_i(t_j, p_i = 1) = \Phi_i(t_{j-1}) \times ((\exp\frac{T_{+,i}}{A}) - 1) \quad \text{A12}$$
$$\Delta\Phi_i(t_j, p_i = -1) = \Phi_i(t_{j-1}) \times ((\exp\frac{T_{-,i}}{A}) - 1)$$

This formulation yields three observations: 1.) the event sensitivity threshold scales linearly with the photon flux at the previous time point and exponentially with the camera parameters, 2.) the event sensitivity threshold increases more rapidly for positive events than negative events and 3.) when the photometric signal is subjected to zero mean noise sources, such the shot noise and dark noise augmented formulation in **Eq. A4B**, we expect that the event sensitivity threshold represents the number of photons required to encode an event impulse with 50% accuracy. We further verify this claim in simulation in **Sec. 3.1**.